\documentstyle[colap,ifthen,calc]{article}

\begin{document}

\input psfig

\newcommand{\Syl}{$\sigma$}             
\newcommand{\Mor}{$\mu$}                
\newcommand{\Sylm}{$\sigma_{\mu}$}      
\newcommand{\Sylmm}{$\sigma_{\mu\mu}$}  
\newcommand{\Sylx}{$\sigma_{x}$} 
\newcommand{\kernel}{B:$\Phi$}          
\newcommand{\residue}{B/$\Phi$}         
\newcommand{\ppc}{O:$\Phi$}             
\newcommand{\npc}{O/$\Phi$}             

\newcommand{\composition}{$\circ$}      
\newcommand{\conc}{$^{\frown}$}         
\newcommand{\estr}{$\varepsilon$}   
\newcommand{\func}[2]
   {\mbox{{\sc #1(}#2{\sc )}}}

\newcommand{\lab}{$\langle$} 
\newcommand{\rab}{$\rangle$} 


\newcounter{boxwidth}
\newcounter{boxhight}
\newcounter{mttlmboxwidth}
\newcounter{mttlmboxhight}
\newcounter{autosegboxwidth}
\newcounter{defaulthight}
\newcounter{strlength}
\newcounter{notiers}
\newcounter{picwidth}
\newcounter{pichight}
\newcounter{lift}
\newcounter{linelen}
\newcounter{xdirection}
\newcounter{ydirection}
\newcounter{curx}
\newcounter{cury}
\newcounter{pictopmargin}

\newlength{\tapenamewidth}
\newlength{\templen}

\setcounter{mttlmboxwidth}{12}%
\setcounter{mttlmboxhight}{12}
\setcounter{autosegboxwidth}{10}%
\setcounter{pictopmargin}{2}

\newcounter{fsmboxwidth}  \setcounter{fsmboxwidth}{30}
\newcounter{fsmcolumns}  \setcounter{fsmcolumns}{7}
\newcounter{fsmrows}
\newcounter{fsmcolumnsx}
\newcounter{fsmrowsx}
\newcounter{nostates}
\newcounter{curstate}
\newcounter{fsmradius}

\newcommand{\fsm}[2]
   {
    \setcounter{boxwidth}{\value{fsmboxwidth}}%
    \setcounter{boxhight}{\value{fsmboxwidth}}%
    \setcounter{nostates}{0}%
    \countstates#1|| END%
    \setcounter{fsmrows}{\value{nostates}/\value{fsmcolumns}}%
    \setcounter{temp}{\value{fsmrows}*\value{fsmcolumns}}%
    \ifthenelse{\value{nostates} = \value{temp}}%
               {}%
               {\stepcounter{fsmrows}}%
    \setcounter{fsmcolumnsx}{\value{fsmcolumns}-1}%
    \setcounter{fsmrowsx}{\value{fsmrows}-1}%
    \setcounter{fsmradius}{\value{boxwidth}-\value{boxwidth}/10}%
    \setcounter{picwidth}{\value{boxwidth}*\value{fsmcolumns}+%
                          \value{boxwidth}*\value{fsmcolumnsx}+%
                          \value{boxwidth}*2}%
    \setcounter{pichight}{\value{boxhight}*\value{fsmrows}+%
                          \value{boxhight}*\value{fsmrowsx}+%
                          \value{boxhight}*2}%
    \setcounter{lift}{\value{pichight}/-2}%
    \rule[\the\value{lift}pt]{3 pt}{\the\value{pichight}pt}%
    \begin{picture}(\the\value{picwidth},0)(0,-\the\value{lift})%
       \setcounter{curstate}{0}%
       \setcounter{curx}{\value{boxhight}}%
       \setcounter{cury}{\value{boxhight}*\value{fsmrowsx}*2+\value{boxhight}}%
       \drawstates#1|| END%
    \end{picture}
   }

\def\drawstates#1,#2,#3|#4 END
   {
    \put(\the\value{curx},\the\value{cury})%
         {\makebox(\the\value{boxwidth},\the\value{boxhight}){#1}}%
    \setcounter{curx}{\value{curx}+\value{boxwidth}/2}%
    \setcounter{cury}{\value{cury}+\value{boxhight}/2}%
    \put(\the\value{curx},\the\value{cury})%
         {\circle{\the\value{boxwidth}}}%
    \ifthenelse{\equal{#3}{y}}%
       {\put(\the\value{curx},\the\value{cury})%
            {\circle{\the\value{fsmradius}}}}%
       {}%
    \setcounter{curx}{\value{curx}-\value{boxwidth}/2}%
    \setcounter{cury}{\value{cury}-\value{boxhight}/2}%
    \stepcounter{curstate}%
    \ifthenelse{\value{curstate} = \value{fsmcolumns}}
               {\setcounter{curx}{\value{boxhight}}%
                \setcounter{cury}{\value{cury}-\value{boxhight}*2}%
                \setcounter{curstate}{0}}%
               {\setcounter{curx}{\value{curx}+\value{boxwidth}*2}}%
    \ifthenelse{\equal{#4}{|}}{}%
               {\drawstates#4 END}%
   }

\def\countstates#1|#2 END
   {\stepcounter{nostates}%
    \ifthenelse{\equal{#2}{|}}{}%
               {\countstates#2 END}%
   }


\newcommand{\mttlmsetwidth}[1]%
   {\setcounter{mttlmboxwidth}{#1}}

\newcommand{\mttlmsethight}[1]%
   {\setcounter{mttlmboxhight}{#1}}

\newcommand{\mttlm}[3]
   {\immediate\write16{MTTLM = #1}%
    \setcounter{boxwidth}{\value{mttlmboxwidth}}%
    \setcounter{boxhight}{\value{mttlmboxhight}}%
    \setcounter{strlength}{0}%
    \setcounter{notiers}{0}%
    \setlength{\tapenamewidth}{0pt}%
    \counttiers#3|| END%
    \countsurfacewidth#1 END%
    \findtapenamewidth#3|| END%
    \setcounter{picwidth}{\value{boxwidth}*\value{strlength}}%
    \setcounter{pichight}{\value{boxhight}*\value{notiers}+%
                          2*\value{boxhight}+\value{pictopmargin}}%
    \setcounter{lift}{\value{pichight}/-2}%
    \rule[\the\value{lift}pt]{0 pt}{\the\value{pichight}pt}%
    \begin{picture}(\the\value{picwidth},0)(0,-\the\value{lift})%
       \let\boxtype=\makebox%
       \setcounter{curx}{0}%
       \setcounter{cury}{0}%
       \displayonetape#1 END%
       \setcounter{curx}{0}%
       \setcounter{cury}{\value{boxhight}}%
       \displaystrings#2|| END%
       \setcounter{curx}{0}%
       \setcounter{cury}{2*\value{boxhight}}%
       \displaymanytapes#3|| END%
    \end{picture}%
    \hspace{\the\tapenamewidth}%
   }

\def\countsurfacewidth#1:#2 END%
   {\countwidth#1| END%
    \maxtapenamewidth#2 END%
   }

\def\displayonetape#1:#2 END%
   {
    \put(\the\value{curx},\the\value{cury})%
         {\framebox(\the\value{picwidth},\the\value{boxhight}){}}%
    \partition#1| END%
    \put(\the\value{curx},\the\value{cury})%
         {\makebox(\the\value{boxwidth},\the\value{boxhight})[l]{\ {\em #2}}}%
    \setcounter{curx}{0}%
    \displaystrings#1|| END%
   }    

\def\displaymanytapes#1|#2 END%
   {\displayonetape#1 END%
    \addtocounter{cury}{-\value{boxhight}}%
    \ifthenelse{\equal{#2}{|}}{}%
               {\displaymanytapes#2 END}%
   }

\def\partition#1#2 END%
   {\addtocounter{curx}{\value{boxwidth}}%
    \ifthenelse{\equal{#1}{-}}{}%
                      {\put(\value{curx},\value{cury})%
                         {\dashbox{0.5}(0,\value{boxhight}){}}}
    \ifthenelse{\equal{#2}{|}}{}%
               {\partition#2 END}%
   }

\def\findtapenamewidth#1:#2|#3 END
   {\maxtapenamewidth#2 END%
    \ifthenelse{\equal{#3}{|}}{}%
               {\findtapenamewidth#3 END}%
   }

\def\maxtapenamewidth#1 END%
   {\settowidth{\templen}{{\em #1}}%
    \ifthenelse{\templen > \tapenamewidth}%
               {\settowidth{\tapenamewidth}{\ {\em #1}}}%
               {}%
   }


\newcommand{\autosegsetwidth}[1]%
   {\setcounter{autosegboxwidth}{#1}}

\newcommand{\autoseg}[2]%
   {\immediate\write16{Autseg Tiers #1}%
    \setcounter{boxwidth}{\value{autosegboxwidth}}%
    \setcounter{boxhight}{\value{autosegboxwidth}}%
    \setcounter{strlength}{0}%
    \setcounter{notiers}{0}%
    \counttiers#1|| END%
    \countwidth#1 END%
    \setcounter{picwidth}{\value{boxwidth}*\value{strlength}}%
    \setcounter{pichight}{2*\value{boxwidth}*\value{notiers}-%
                          \value{boxwidth}+\value{pictopmargin}}%
    \setcounter{lift}{\value{pichight}/-2}%
    \rule[\the\value{lift}pt]{0 pt}{\the\value{pichight}pt}%
    \begin{picture}(\the\value{picwidth},0)(0,-\the\value{lift})%
       \let\boxtype=\makebox%
       \setcounter{curx}{0}%
       \setcounter{cury}{0}%
       \displaystrings#1|| END%
       \setcounter{curx}{\value{boxwidth}/2}%
       \linkstrings#2,||| END%
    \end{picture}%
   }

\def\counttiers#1|#2 END
   {\stepcounter{notiers}%
    \ifthenelse{\equal{#2}{|}}{}%
               {\counttiers#2 END}%
   }

\def\countwidth#1|#2 END%
   {\countlength#1| END%
   }

\def\countlength#1#2 END
   {\stepcounter{strlength}%
    \ifthenelse{\equal{#2}{|}}{}%
               {\countlength#2 END}%
   }

\def\displaystrings#1|#2 END
   {\setmorpheme#1| END%
    \setcounter{curx}{0}%
    \addtocounter{cury}{2*\value{boxhight}}%
    \ifthenelse{\equal{#2}{|}}{}%
               {\displaystrings#2 END}%
   }

\def\setmorpheme#1#2 END%
   {\ifthenelse{\equal{#1}{-}}{}%
               {\put(\the\value{curx},\the\value{cury})%
                {\boxtype(\the\value{boxwidth},\the\value{boxhight}){#1}}}%
    \addtocounter{curx}{\value{boxwidth}}%
    \ifthenelse{\equal{#2}{|}}{}%
               {\setmorpheme#2 END}%
   }

\def\linkstrings#1#2#3,#4 END
   {
    \ifthenelse{\equal{#1}{-}}{}%
       {
        \ifthenelse{#1 < #2}%
               {\setcounter{cury}{(2*#1-1)*\value{boxwidth}}%
                \setcounter{ydirection}{1}%
                \setcounter{linelen}{(2*(#2-#1)-1)*\value{boxwidth}}}%
               {}%
        \ifthenelse{#1 > #2}%
               {\setcounter{cury}{2*(#1-1)*\value{boxwidth}}%
                \setcounter{ydirection}{-1}%
                \setcounter{linelen}{(2*(#1-#2)-1)*\value{boxwidth}}}%
               {}%
        \setcounter{xdirection}{0}%
        \ifthenelse{\equal{#3}{r}}{\setcounter{xdirection}{1}}{}%
        \ifthenelse{\equal{#3}{rr}}{\setcounter{xdirection}{2}%
                                   \addtocounter{linelen}{\value{boxwidth}}}{}%
        \ifthenelse{\equal{#3}{rrr}}{\setcounter{xdirection}{3}%
                                   \addtocounter{linelen}{2*\value{boxwidth}}}%
                                   {}%
        \ifthenelse{\equal{#3}{l}}{\setcounter{xdirection}{-1}}{}%
        \ifthenelse{\equal{#3}{ll}}{\setcounter{xdirection}{-2}%
                                   \addtocounter{linelen}{\value{boxwidth}}}{}%
        \ifthenelse{\equal{#3}{lll}}{\setcounter{xdirection}{-3}%
                                   \addtocounter{linelen}{2*\value{boxwidth}}}%
                                   {}%
        \put(\value{curx},\value{cury})%
            {\line(\value{xdirection},\value{ydirection})%
            {\value{linelen}}}}%
    \addtocounter{curx}{\value{boxwidth}}%
    \ifthenelse{\equal{#4}{|||}}{}%
               {\linkstrings#4 END}%
   }

\newcounter{moraboxwidth}
\setcounter{moraboxwidth}{10}

\newcommand{\morasetwidth}[1]%
   {\setcounter{moraboxwidth}{#1}}

\newcommand{\moratree}[2]%
   {
    \setcounter{picwidth}{2*\value{moraboxwidth}}%
    \setcounter{pichight}{4*\value{moraboxwidth}+\value{moraboxwidth}/2}%
    \setcounter{lift}{\value{pichight}/-2}%
    \rule[\the\value{lift}pt]{0 pt}{\the\value{pichight}pt}%
    \begin{picture}(\the\value{picwidth},0)(0,-\the\value{lift})%
       \drawbasic{#1}{#2}%
    \end{picture}%
   }

\newcommand{\mmoratree}[3]%
   {
    \setcounter{picwidth}{3*\value{moraboxwidth}}%
    \setcounter{pichight}{4*\value{moraboxwidth}+\value{moraboxwidth}/2}%
    \setcounter{lift}{\value{pichight}/-2}%
    \rule[\the\value{lift}pt]{0 pt}{\the\value{pichight}pt}%
    \begin{picture}(\the\value{picwidth},0)(0,-\the\value{lift})%
       \drawbasic{#1}{#2}%
       \setcounter{curx}{2*\value{moraboxwidth}}%
       \put(\value{curx},0)%
          {\makebox(\value{moraboxwidth},\value{moraboxwidth})[b]{#3}}%
       \setcounter{cury}{2*\value{moraboxwidth}}%
       \put(\value{curx},\value{cury})%
          {\makebox(\value{moraboxwidth},\value{moraboxwidth}){\Mor}}%
       \setcounter{curx}{2*\value{moraboxwidth}+\value{moraboxwidth}/2}%
       \put(\value{curx},\value{cury}){\line(0,-1){\value{moraboxwidth}}}%
       \setcounter{curx}{\value{moraboxwidth}+\value{moraboxwidth}/2}%
       \setcounter{cury}{3*\value{moraboxwidth}+\value{moraboxwidth}/2}%
       \put(\value{curx},\value{cury}){\line(2,-1){\value{moraboxwidth}}}%
    \end{picture}%
   }

\newcommand{\xmoratree}[1]%
   {
    \setcounter{picwidth}{\value{moraboxwidth}}%
    \setcounter{pichight}{4*\value{moraboxwidth}+\value{moraboxwidth}/2}%
    \setcounter{lift}{\value{pichight}/-2}%
    \rule[\the\value{lift}pt]{0 pt}{\the\value{pichight}pt}%
    \begin{picture}(\the\value{picwidth},0)(0,-\the\value{lift})%
       \put(0,0){\makebox(\value{moraboxwidth},\value{moraboxwidth})[b]{#1}}%
       \setcounter{cury}{3*\value{moraboxwidth}+\value{moraboxwidth}/2}%
       \put(0,\value{cury})%
       {\makebox(\value{moraboxwidth},\value{moraboxwidth}){\Sylx}}%
       \setcounter{curx}{\value{moraboxwidth}/2}%
       \setcounter{linelen}{2*\value{moraboxwidth}+\value{moraboxwidth}/2}%
       \put(\value{curx},\value{cury}){\line(0,-1){\value{linelen}}}%
    \end{picture}%
   }

\newcommand{\gmoratree}[5]%
   {\mbox{\mmoratree{#1}{#2}{#3}%
          \hspace{-\value{moraboxwidth}pt}%
          \mmoratree{{}}{#4}{#5}}%
   \immediate\write16{(#1,#2,#3,#4,#5)}%
   }

\newcommand{\drawbasic}[2]%
   {
    \put(0,0){\makebox(\value{moraboxwidth},\value{moraboxwidth})[b]{#1}}%
    \put(\value{moraboxwidth},0)%
       {\makebox(\value{moraboxwidth},\value{moraboxwidth})[b]{#2}}%
    \setcounter{cury}{2*\value{moraboxwidth}}%
    \put(\value{moraboxwidth},\value{cury})%
       {\makebox(\value{moraboxwidth},\value{moraboxwidth}){\Mor}}%
    \setcounter{cury}{3*\value{moraboxwidth}+\value{moraboxwidth}/2}%
    \put(\value{moraboxwidth},\value{cury})%
       {\makebox(\value{moraboxwidth},\value{moraboxwidth}){\Syl}}%
    \setcounter{curx}{\value{moraboxwidth}+\value{moraboxwidth}/2}%
    \setcounter{cury}{3*\value{moraboxwidth}+\value{moraboxwidth}/2}%
    \put(\value{curx},\value{cury}){\line(-2,-5){\value{moraboxwidth}}}%
    \setcounter{linelen}{\value{moraboxwidth}/2}%
    \put(\value{curx},\value{cury}){\line(0,-1){\value{linelen}}}%
    \setcounter{cury}{2*\value{moraboxwidth}}%
    \put(\value{curx},\value{cury}){\line(0,-1){\value{moraboxwidth}}}%
   }

\newcounter{tapeboxhight}
\setcounter{tapeboxhight}{15}
\newcounter{delta}
\newcounter{fstwidth}
\newcounter{temp}

\newcommand{\tapehight}[1]%
   {\setcounter{tapeboxhight}{#1}}

\newcommand{\cascadetransducers}%
   {
    \setcounter{picwidth}{10*\value{tapeboxhight}}%
    \setcounter{pichight}{9*\value{tapeboxhight}}%
    \setcounter{lift}{\value{pichight}/-2}%
    \rule[\the\value{lift}pt]{0 pt}{\the\value{pichight}pt}%
    \begin{picture}(\the\value{picwidth},0)(0,-\the\value{lift})%
       \put(0,0){\framebox(\value{picwidth},\value{tapeboxhight})
             {Surface String}}%
       \setcounter{cury}{4*\value{tapeboxhight}}
       \setcounter{temp}{8*\value{tapeboxhight}}
       \put(\value{tapeboxhight},\value{cury})
             {\framebox(\value{temp},\value{tapeboxhight})
             {Intermediate String}}%
       \setcounter{cury}{8*\value{tapeboxhight}}
       \put(0,\value{cury}){\framebox(\value{picwidth},\value{tapeboxhight})
             {Lexical String}}%
       \setcounter{curx}{\value{picwidth}/2}
       \setcounter{cury}{\value{tapeboxhight}}
       \setcounter{delta}{2*\value{tapeboxhight}}
       \multiput(\value{curx},\value{cury})(0,\value{delta}){4}
          {\line(0,1){\value{tapeboxhight}}}
       \setcounter{cury}{2*\value{tapeboxhight}+\value{tapeboxhight}/2}
       \setcounter{delta}{4*\value{tapeboxhight}}
       \setcounter{fstwidth}{2*\value{tapeboxhight}}
       \multiput(\value{curx},\value{cury})(0,\value{delta}){2}
          {\oval(\value{fstwidth},\value{tapeboxhight})}
       \put(\value{curx},\value{cury}){\makebox(0,0){\em FST$_n$}}%
       \setcounter{cury}{6*\value{tapeboxhight}+\value{tapeboxhight}/2}
       \put(\value{curx},\value{cury}){\makebox(0,0){\em FST$_1$}}%
    \end{picture}%
    ~$\Longrightarrow$~%
    \begin{picture}(\the\value{picwidth},0)(0,-\the\value{lift})%
       \put(0,0){\framebox(\value{picwidth},\value{tapeboxhight})
             {Surface String}}%
       \setcounter{cury}{8*\value{tapeboxhight}}
       \put(0,\value{cury}){\framebox(\value{picwidth},\value{tapeboxhight})
             {Lexical String}}%
       \setcounter{curx}{\value{picwidth}/2}
       \setcounter{cury}{\value{tapeboxhight}}
       \setcounter{temp}{3*\value{tapeboxhight}}
       \setcounter{delta}{4*\value{tapeboxhight}}
       \multiput(\value{curx},\value{cury})(0,\value{delta}){2}
          {\line(0,1){\value{temp}}}
       \setcounter{cury}{4*\value{tapeboxhight}+\value{tapeboxhight}/2}
       \setcounter{fstwidth}{8*\value{tapeboxhight}}
       \put(\value{curx},\value{cury})
              {\oval(\value{fstwidth},\value{tapeboxhight})}
       \put(\value{curx},\value{cury})
              {\makebox(0,0){$FST_1 \circ FST_2 \circ \cdots \circ FST_n$}}%
    \end{picture}%
   }

\newcommand{\paralleltransducers}%
   {\paralleltransducersone%
    ~$\Longrightarrow$~%
    \paralleltransducerstwo%
   }

\newcommand{\paralleltransducersone}%
   {
    \setcounter{picwidth}{10*\value{tapeboxhight}}%
    \setcounter{pichight}{9*\value{tapeboxhight}}%
    \setcounter{lift}{\value{pichight}/-2}%
    \rule[\the\value{lift}pt]{0 pt}{\the\value{pichight}pt}%
    \begin{picture}(\the\value{picwidth},0)(0,-\the\value{lift})%
       \put(0,0){\framebox(\value{picwidth},\value{tapeboxhight})
             {Surface String}}%
       \setcounter{cury}{8*\value{tapeboxhight}}
       \put(0,\value{cury}){\framebox(\value{picwidth},\value{tapeboxhight})
             {Lexical String}}%
       \setcounter{curx}{\value{picwidth}/2}
       \setcounter{cury}{\value{tapeboxhight}}
       \setcounter{delta}{6*\value{tapeboxhight}}
       \multiput(\value{curx},\value{cury})(0,\value{delta}){2}
          {\line(0,1){\value{tapeboxhight}}}
       \setcounter{curx}{\value{tapeboxhight}}
       \setcounter{cury}{2*\value{tapeboxhight}}
       \setcounter{delta}{5*\value{tapeboxhight}}
       \setcounter{temp}{8*\value{tapeboxhight}}
       \multiput(\value{curx},\value{cury})(0,\value{delta}){2}
          {\line(1,0){\value{temp}}}
       \setcounter{delta}{3*\value{tapeboxhight}}
       \setcounter{temp}{2*\value{tapeboxhight}}
       \multiput(\value{curx},\value{cury})(\value{delta},0){2}
          {\line(0,1){\value{temp}}}
       \setcounter{cury}{5*\value{tapeboxhight}}
       \multiput(\value{curx},\value{cury})(\value{delta},0){2}
          {\line(0,1){\value{temp}}}
       \setcounter{cury}{4*\value{tapeboxhight}+\value{tapeboxhight}/2}
       \setcounter{fstwidth}{2*\value{tapeboxhight}}
       \multiput(\value{curx},\value{cury})(\value{delta},0){2}
          {\oval(\value{fstwidth},\value{tapeboxhight})}
       \put(\value{curx},\value{cury}){\makebox(0,0){\em FST$_1$}}%
       \setcounter{curx}{4*\value{tapeboxhight}}
       \put(\value{curx},\value{cury}){\makebox(0,0){\em FST$_2$}}%
       \setcounter{curx}{9*\value{tapeboxhight}}
       \setcounter{cury}{2*\value{tapeboxhight}}
       \setcounter{delta}{3*\value{tapeboxhight}}
       \multiput(\value{curx},\value{cury})(0,\value{delta}){2}
          {\line(0,1){\value{temp}}}
       \setcounter{cury}{4*\value{tapeboxhight}+\value{tapeboxhight}/2}
       \put(\value{curx},\value{cury})
          {\oval(\value{fstwidth},\value{tapeboxhight})}
       \put(\value{curx},\value{cury}){\makebox(0,0){\em FST$_n$}}%
       \setcounter{curx}{6*\value{tapeboxhight}+\value{tapeboxhight}/2}
       \put(\value{curx},\value{cury}){\makebox(0,0){$\cdots$}}%
    \end{picture}%
   }
\newcommand{\paralleltransducerstwo}%
   {\begin{picture}(\the\value{picwidth},0)(0,-\the\value{lift})%
       \put(0,0){\framebox(\value{picwidth},\value{tapeboxhight})
             {Surface String}}%
       \setcounter{cury}{8*\value{tapeboxhight}}
       \put(0,\value{cury}){\framebox(\value{picwidth},\value{tapeboxhight})
             {Lexical String}}%
       \setcounter{curx}{\value{picwidth}/2}
       \setcounter{cury}{\value{tapeboxhight}}
       \setcounter{temp}{3*\value{tapeboxhight}}
       \setcounter{delta}{4*\value{tapeboxhight}}
       \multiput(\value{curx},\value{cury})(0,\value{delta}){2}
          {\line(0,1){\value{temp}}}
       \setcounter{cury}{4*\value{tapeboxhight}+\value{tapeboxhight}/2}
       \setcounter{fstwidth}{8*\value{tapeboxhight}}
       \put(\value{curx},\value{cury})
              {\oval(\value{fstwidth},\value{tapeboxhight})}
       \put(\value{curx},\value{cury})
              {\makebox(0,0){$FST_1 \cap FST_2 \cap \cdots \cap FST_n$}}%
    \end{picture}%
   }

\newcommand{\uniontransducers}%
   {
    \setcounter{picwidth}{10*\value{tapeboxhight}}%
    \setcounter{pichight}{9*\value{tapeboxhight}}%
    \setcounter{lift}{\value{pichight}/-2}%
    \rule[\the\value{lift}pt]{0 pt}{\the\value{pichight}pt}%
    \begin{picture}(\the\value{picwidth},0)(0,-\the\value{lift})%
       \put(0,0){\framebox(\value{picwidth},\value{tapeboxhight})
             {Surface String}}%
       \setcounter{curx}{2*\value{tapeboxhight}+\value{tapeboxhight}/2}
       \setcounter{delta}{2*\value{tapeboxhight}+\value{tapeboxhight}/2}
       \multiput(\value{curx},0)(\value{delta},0){3}
          {\dashbox{.75}(0,\value{tapeboxhight}){}}
       \setcounter{cury}{8*\value{tapeboxhight}}
       \put(0,\value{cury}){\framebox(\value{picwidth},\value{tapeboxhight})
             {Lexical String}}%
       \setcounter{cury}{8*\value{tapeboxhight}}
       \multiput(\value{curx},\value{cury})(\value{delta},0){3}
          {\dashbox{.75}(0,\value{tapeboxhight}){}}
       \setcounter{curx}{\value{tapeboxhight}}
       \setcounter{cury}{\value{tapeboxhight}}
       \setcounter{delta}{3*\value{tapeboxhight}}
       \multiput(\value{curx},\value{cury})(\value{delta},0){2}
          {\line(0,1){\value{delta}}}
       \setcounter{cury}{5*\value{tapeboxhight}}
       \multiput(\value{curx},\value{cury})(\value{delta},0){2}
          {\line(0,1){\value{delta}}}
       \setcounter{cury}{4*\value{tapeboxhight}+\value{tapeboxhight}/2}
       \setcounter{fstwidth}{2*\value{tapeboxhight}}
       \multiput(\value{curx},\value{cury})(\value{delta},0){2}
          {\oval(\value{fstwidth},\value{tapeboxhight})}
       \put(\value{curx},\value{cury}){\makebox(0,0){\em FST$_1$}}%
       \setcounter{curx}{4*\value{tapeboxhight}}
       \put(\value{curx},\value{cury}){\makebox(0,0){\em FST$_2$}}%
       \setcounter{curx}{9*\value{tapeboxhight}}
       \setcounter{cury}{\value{tapeboxhight}}
       \setcounter{delta}{4*\value{tapeboxhight}}
       \multiput(\value{curx},\value{cury})(0,\value{delta}){2}
          {\line(0,1){\value{temp}}}
       \setcounter{cury}{4*\value{tapeboxhight}+\value{tapeboxhight}/2}
       \put(\value{curx},\value{cury})
          {\oval(\value{fstwidth},\value{tapeboxhight})}
       \put(\value{curx},\value{cury}){\makebox(0,0){\em FST$_n$}}%
       \setcounter{curx}{6*\value{tapeboxhight}+\value{tapeboxhight}/2}
       \put(\value{curx},\value{cury}){\makebox(0,0){$\cdots$}}%
    \end{picture}%
    ~$\Longrightarrow$~%
    \begin{picture}(\the\value{picwidth},0)(0,-\the\value{lift})%
       \put(0,0){\framebox(\value{picwidth},\value{tapeboxhight})
             {Surface String}}%
       \setcounter{curx}{2*\value{tapeboxhight}+\value{tapeboxhight}/2}
       \setcounter{delta}{2*\value{tapeboxhight}+\value{tapeboxhight}/2}
       \multiput(\value{curx},0)(\value{delta},0){3}
          {\dashbox{.75}(0,\value{tapeboxhight}){}}
       \setcounter{cury}{8*\value{tapeboxhight}}
       \put(0,\value{cury}){\framebox(\value{picwidth},\value{tapeboxhight})
             {Lexical String}}%
       \setcounter{cury}{8*\value{tapeboxhight}}
       \multiput(\value{curx},\value{cury})(\value{delta},0){3}
          {\dashbox{.75}(0,\value{tapeboxhight}){}}
       \setcounter{curx}{\value{picwidth}/2}
       \setcounter{cury}{4*\value{tapeboxhight}+\value{tapeboxhight}/2}
       \setcounter{fstwidth}{8*\value{tapeboxhight}}
       \put(\value{curx},\value{cury})
              {\oval(\value{fstwidth},\value{tapeboxhight})}
       \put(\value{curx},\value{cury})
              {\makebox(0,0){$FST_1 \cup FST_2 \cup \cdots \cup FST_n$}}%
       \setcounter{cury}{5*\value{tapeboxhight}}
       \setcounter{temp}{4*\value{tapeboxhight}}
       \put(\value{curx},\value{cury}){\line(-4,3){\value{temp}}}
       \put(\value{curx},\value{cury}){\line(4,3){\value{temp}}}
       \put(\value{curx},\value{cury}){\line(-1,3){\value{tapeboxhight}}}
       \put(\value{curx},\value{cury}){\line(1,3){\value{tapeboxhight}}}
       \setcounter{cury}{4*\value{tapeboxhight}}
       \setcounter{temp}{4*\value{tapeboxhight}}
       \put(\value{curx},\value{cury}){\line(-4,-3){\value{temp}}}
       \put(\value{curx},\value{cury}){\line(4,-3){\value{temp}}}
       \put(\value{curx},\value{cury}){\line(-1,-3){\value{tapeboxhight}}}
       \put(\value{curx},\value{cury}){\line(1,-3){\value{tapeboxhight}}}
    \end{picture}%
   }

\newcommand{\katajakoskenniemi}%
   {
    \setcounter{picwidth}{10*\value{tapeboxhight}}%
    \setcounter{pichight}{9*\value{tapeboxhight}}%
    \setcounter{lift}{\value{pichight}/-2}%
    \rule[\the\value{lift}pt]{0 pt}{\the\value{pichight}pt}%
    \begin{picture}(\the\value{picwidth},0)(0,-\the\value{lift})%
       \put(0,0){\makebox(\value{picwidth},\value{tapeboxhight})
             {Surface Representation}}%
       \setcounter{cury}{4*\value{tapeboxhight}}
       \setcounter{temp}{8*\value{tapeboxhight}}
       \put(\value{tapeboxhight},\value{cury})
             {\makebox(\value{temp},\value{tapeboxhight})
             {Lexical Representation}}%
       \setcounter{cury}{8*\value{tapeboxhight}}
       \put(0,\value{cury}){\makebox(\value{picwidth},\value{tapeboxhight})
             {Lexical Entries (Morphemes)}}%
       \setcounter{curx}{\value{picwidth}/2}
       \setcounter{cury}{\value{tapeboxhight}}
       \setcounter{delta}{2*\value{tapeboxhight}}
       \multiput(\value{curx},\value{cury})(0,\value{delta}){4}
          {\line(0,1){\value{tapeboxhight}}}
       \setcounter{cury}{2*\value{tapeboxhight}+\value{tapeboxhight}/2}
       \setcounter{delta}{4*\value{tapeboxhight}}
       \setcounter{fstwidth}{10*\value{tapeboxhight}}
       \multiput(\value{curx},\value{cury})(0,\value{delta}){2}
          {\oval(\value{fstwidth},\value{tapeboxhight})}
       \put(\value{curx},\value{cury}){\makebox(0,0){\sc Two-Level Rules}}%
       \setcounter{cury}{6*\value{tapeboxhight}+\value{tapeboxhight}/2}
       \put(\value{curx},\value{cury}){\makebox(0,0){\sc Lexicon Component}}%
    \end{picture}%
   }

\newcommand{\environbar}{\underline{\hspace*{1.5em}}\ }

\newcommand{\phonrule}[4]%
   {#1 {}$\rightarrow${} #2 / #3 \environbar #4}

\newcommand{\tlr}[7]%
   {\begin{tabular}{cccccc}%
      {}#5&--&#6&--&#7&#4 \\
      {}#1&--&#2&--&#3&
   \end{tabular}%
   \vspace{.1in}}

\newcommand{\tlrf}[8]%
   {\begin{tabular}{cccccc}%
      {}#5&--&#6&--&#7&#4\\
      {}#1&--&#2&--&#3\\
      \multicolumn{6}{l}{{\sf Features:} {\tt #8}}
   \end{tabular}
   \vspace{.1in}}

\newcommand{\tlrc}[8]%
   {\begin{tabular}{cccccc}%
      {}#5&--&#6&--&#7&#4 \\
      {}#1&--&#2&--&#3& \\
      \multicolumn{6}{l}{{\sf where} #8}%
   \end{tabular}%
   \vspace{.1in}}

\newcommand{\tlrt}[8]%
   {\tlrc#1#2#3#4#5#6#7#8}

\newcommand{\tlrule}[9]
   {\begin{tabbing}%
       tl\_rule({\tt #1},
                  \= {\tt #2}, \= {\tt #3}, \= {\tt #4}, {\tt #5},\\%
                  \> {\tt #6}, \> {\tt #7}, \> {\tt #8},\\%
                  \> \restoftlrule#9END%
    \end{tabbing}%
   }

\newcommand{\tlruleintab}[9]
   {tl\_rule({\tt #1},
                  \= {\tt #2}, \= {\tt #3}, \= {\tt #4}, {\tt #5},\\%
             \>   \> {\tt #6}, \> {\tt #7}, \> {\tt #8},\\%
             \>   \> \restoftlrule#9END%
   }

\newcommand{\tlruleintablong}[9]
   {tl\_rule({\tt #1},
                  \= {\tt #2}, {\tt #3}, {\tt #4}, {\tt #5},
                     {\tt #6}, {\tt #7}, {\tt #8},\\%
              \>  \> \restoftlrule#9END%
   }

\def\restoftlrule#1|#2END%
  {{\tt #1}, {\tt #2}).}

%

\newcounter{examplectr}
\newcounter{subexamplectr}
%
\newenvironment{ex}%
   {\addtocounter{examplectr}{1}%
     \setcounter{subexamplectr}{0}%
     \begin{list}{(\arabic{examplectr})}%
                 {\setlength{\topsep}{.1in}%
           	  \setlength{\leftmargin}{0.45in}%
	          \setlength{\labelsep}{0.075in}}%
     \item \begin{minipage}[t]{5.5in}%
   }%
   {\end{minipage}%
    \end{list}\vspace{.1in}}%
%
\newenvironment{subex}%
   { \addtocounter{subexamplectr}{1}
     \begin{list}
       {\alph{subexamplectr}.}%
       {\setlength{\topsep}{-\parskip}
	\setlength{\leftmargin}{0.175in}
	\setlength{\labelsep}{0.075in}}
       \item
   }%
   {\end{list}}
%
\newcommand{\exnum}[2]{\addtocounter{examplectr}{#1}(\arabic{examplectr}{#2})\addtocounter{examplectr}{-#1}}


\newtheorem{loctheorem}{Theorem}[section]
\newtheorem{locexample}{Example}[section]
\newtheorem{locdef}{Definition}[section]
\newtheorem{loclemma}{Lemma}[section]

\newenvironment{theorem}{\begin{loctheorem}\em }{\end{loctheorem}}
\newenvironment{example}{\begin{locexample}\em }{\hfill$\Box$\end{locexample}}
\newenvironment{definition}{\begin{locdef}\em }{\hfill$\Box$\end{locdef}}
\newenvironment{lemma}{\begin{loclemma}\em }{\hfill$\Box$\end{loclemma}}
\newenvironment{proof}{{\em Proof \ }}{\hfill$\Box$}

\newcommand{\disjprod}{\bar{\cap}}
\newcommand{\ifpthens}[2]
   {\overline{#1\ \overline{#2}}}
\newcommand{\ifsthenp}[2]
   {\overline{\overline{#1}\ #2}}


\author{Tanya Bowden \thanks{}
\and George Anton Kiraz \thanks{}\\
 University of Cambridge \\
 Computer Laboratory \\
 Pembroke Street, Cambridge CB2 3QG \\
 {\tt \{Tanya.Bowden, George.Kiraz\}@cl.cam.ac.uk} \\
 {\tt http://www.cl.cam.ac.uk/users/\{tgb1000, gk105\}} \\
  \mbox{}}

\author{\parbox[t]{.333\linewidth}{\centering Edmund Grimley-Evans\thanks{\ Supported by SERC studentship no.~92313384.} \\
          \small University of Cambridge \\ (St John's College)
\\ Computer Laboratory \\ Cambridge CB2 3QG, UK
\\ {\tt Edmund.Grimley-Evans@cl.cam.ac.uk}}
        \parbox[t]{.333\linewidth}{\centering George Anton Kiraz\thanks{\ Supported by a Benefactors' Studentship
from St John's College.} \\
          \small University of Cambridge \\ (St John's College)
\\ Computer Laboratory \\ Cambridge CB2 3QG, UK
\\ {\tt George.Kiraz@cl.cam.ac.uk}}
       \parbox[t]{.333\linewidth}{\centering Stephen G. Pulman \\
          \small University of Cambridge
\\ Computer Laboratory \\ Cambridge CB2 3QG, UK
\\ and SRI International, Cambridge
\\ {\tt sgp@cam.sri.com}}}



\title{Compiling a Partition-Based Two-Level Formalism}

\maketitle

\begin{abstract}
This paper describes an algorithm for the compilation of a two (or
more) level orthographic or phonological rule notation into finite
state transducers. The notation is an alternative to the standard one
deriving from Koskenniemi's work: it is believed to have some
practical descriptive advantages, and is quite widely used, but has a
different interpretation. Efficient interpreters exist for the
notation, but until now it has not been clear how to compile to
equivalent automata in a transparent way. The present paper shows how
to do this, using some of the conceptual tools provided by Kaplan and
Kay's regular relations calculus.
\end{abstract}

\section{Introduction}
\label{intro}

\newcommand{\IT}{{partition \,}}

Two-level formalisms based on that introduced by \cite{Koskenniemi:83}
(see also \cite{Ritchie:92book} and \cite{Kaplan:94}) are widely used in
practical NLP
systems, and are deservedly
regarded as something of a standard. However, there is at least one
serious rival two-level notation in existence, developed in response
to practical difficulties encountered in writing large-scale
morphological descriptions using Koskenniemi's notation.  The formalism
was first introduced in \cite{Black:87}, was adapted by
\cite{Ruessink:89}, and an extended version of it was proposed for use in
the European Commission's ALEP language engineering platform
\cite{Pulman:91}. A further extension to the formalism was described
in \cite{Pulman:93}. 

The alternative \IT formalism was motivated by several perceived practical
disadvantages to Koskenniemi's notation. These are detailed more fully
in \cite[pp.~13-15]{Black:87}, and in
\cite[pp.~181-9]{Ritchie:92book}. In brief:
(1) Koskenniemi rules are not easily interpretable (by the
grammarian) locally, for the interpretation of `feasible pairs'
depends on 
other rules in the set.
(2) There are frequently interactions between rules: whenever the
lexical/surface 
pair affected by a rule A appears in the context of another rule B,
the grammarian must check that its appearance in rule B will not
conflict with the requirements of rule A. 
(3) Contexts may conflict: the same lexical character may
obligatorily have multiple realisations in different contexts, but it
may be impossible to state the contexts in ways that do not block a
desired application.
(4) Restriction to single character changes: whenever a change
affecting more than one adjacent character occurs, multiple
rules must be written. At best this prompts the interaction problem,
and at worst can require the rules to be formulated with 
under-restrictive contexts to avoid mutual blocking.
(5) There is no mechanism for relating particular rules to specific
classes of morpheme. This has to be achieved indirectly by introducing
special abstract triggering characters in lexical representations.
This is clumsy, and 
sometimes
descriptively inadequate \cite{Trost:90}.

Some of these problems can be alleviated by the use of a rule compiler
that detects conflicts such as that described in
\cite{Karttunen:92}. Others could be overcome by simple extensions to
the formalism. But
several of these problems arise from the interpretation of Koskenniemi
rules: each rule corresponds to a transducer, and the two-level
description of a language consists of the intersection of these transducers.
Thus somehow or other it must be arranged that every rule accepts every
two-level correspondence. We refer to this class of formalisms as `parallel':
every rule, in effect, is applied in parallel at each point in the input.

The \IT formalism consists of two types of rules (defined in more detail
below) which enforce optional or obligatory changes. The notion of
well-formedness is defined via the notion of a `partition' of
a sequence of lexical/surface correspondences. Informally,
a partition is a valid analysis if
(i) every element of the partition is licensed by an optional
rule, and
(ii) no element of the partition violates an obligatory rule.

We have found that this formalism has some practical  advantages:
(1) The rules are relatively independent of each other.
(2) Their interpretation is more familiar for linguists: each rule
copes with a single correspondence: in general you don't have to worry about all
other rules having to be compatible with it.
(3) Multiple character changes are permitted (with some restrictions discussed below).
(4) A category or term associated with each rule is required to
unify with the affected morpheme, allowing for morpho-syntactic effects
to be cleanly described.
(5) There is a simple and  efficient  direct interpreter for the rule
formalism.

The \IT formalism has been implemented in the European Commission's
ALEP system for natural language engineering, distributed to over 30
sites.  Descriptions of 9 EU languages are being developed.  A
version has also been implemented within SRI's Core Language Engine
\cite{Carter:95} and has been used to develop descriptions of
English, French, Spanish, Polish, Swedish, and Korean morphology.  An
N-level extension of the formalism has also been developed by
\cite{Kiraz:94Coling,Kiraz:thesis} and used to describe the morphology
of Syriac and other Semitic languages, and by \cite{Bowden:95} for
error detection in nonconcatenative strings.  This partition-based
two-level formalism is thus a serious rival
to the standard Koskenniemi notation.

However, until now, the Koskenniemi notation has had one clear
advantage in that it was clear how to compile it into transducers,
with all the consequent gains in efficiency and portability and
with the ability to construct lexical transducers as in \cite{Karttunen:94}.
This
paper sets out to remedy that defect by describing a compilation
algorithm for the partition-based two-level notation.

\section{Definition of the Formalism}
\label{definition}

\subsection{Formal Definition}

We use $n$ tapes, where the first $N$ tapes are lexical and the
remaining $M$ are surface, $n=N+M$. In practice, $M=1$. We write
$\Sigma_i$ for the alphabet of symbols used on tape $i$, and $\Sigma =
(\Sigma_1\cup\{\epsilon\})\times...\times(\Sigma_n\cup\{\epsilon\})$,
so that $\Sigma^*$ is the set of string-tuples representing possible
contents of the $n$ tapes. A proper subset of regular n-relations
have the property that they are
expressible as the Cartesian product of $n$ regular languages,
$R=R_1\times...\times R_n$; we call such relations `orthogonal'.
(We present our definitions along the lines of \cite{Kaplan:94}).

We use two regular operators: {\tt Intro} and {\tt Sub}. ${\tt
Intro}_S L$ denotes the set of strings in $L$ into which elements of
$S$ may be arbitrarily inserted, and ${\tt Sub}_{A,B} L$ denotes the
set of strings in $L$ in which substrings that are in $B$ may be
replaced by strings from $A$. Both operators map regular languages
into regular languages, because they can be characterised by regular
relations: over the alphabet $\Sigma$, ${\tt Intro}_S = ({\tt
Id}_\Sigma\cup(\{\epsilon\}\times S))^*$, ${\tt Sub}_{A,B} = ({\tt
Id}_\Sigma\cup(B\times A))^*$, where ${\tt Id}_L =
\{(s,s)\mid s\in L\}$, the identity relation over $L$.

\medskip

There are two kinds of two-level rules. The context restriction, or
optional, rules, consist of a left context $l$, a centre $c$, and a
right context $r$. Surface coercion, or obligatory, rules require the
centre to be split into lexical $c_l$ and surface $c_s$ components.

\begin{definition}
   \label{cr}
   A $N$:$M$ {\bf context restriction (CR) rule} is a triple
   $(l,c,r)$ where $l,c,r$ are `orthogonal' regular relations of the
   form $l = l_1\times...\times l_n$, $c = c_1\times...\times c_n$,
   $r = r_1\times...\times r_n$. 
\end{definition}

\begin{definition}
   \label{sc}
   A $N$:$M$ {\bf surface coercion (SC) rule} is a quadruple
   $(l,c_l,c_s,r)$ where $l$ and $r$ are `orthogonal' regular
   relations of the form $l = l_1\times...\times l_n$, $r =
   r_1\times...\times r_n$, and $c_l$ and $ c_s$ are `orthogonal'
   regular relations restricting only the lexical and surface tapes,
   respectively, of the form $c_l = c_1\times...\times
   c_N\times\Sigma_{N+1}^*\times...\times\Sigma_{N+M}^*$ and $c_s =
   \Sigma_1^*\times...\times\Sigma_N^*\times c_{N+1}\times...\times
   c_{N+M}$.
\end{definition}
\medskip

We usually use the following notation for rules:

      \tlr{LSC}{\sc Surf}{RSC}{$\Rightarrow\mid\Leftarrow\mid\Leftrightarrow$}
          {LLC}{\sc Lex}{RLC}\\
where \\
\begin{tabular}{l}
{\sc LLC} (left lexical context) = $\langle l_1,\ldots,l_N\rangle$\\
{\sc Lex} (lexical form) = $\langle c_1,\ldots,c_N\rangle$\\
{\sc RLC} (right lexical context) = $\langle r_1,\ldots,r_N\rangle$\\
{\sc LSC} (left surface context) = $\langle l_{N+1},\ldots,l_{N+M}\rangle$\\
{\sc Surf} (surface form) = $\langle c_{N+1},\ldots,c_{N+M}\rangle$\\
{\sc RSC} (right surface context) = $\langle r_{N+1},\ldots,r_{N+M}\rangle$
\end{tabular}
\smallskip


Since in practice all the left contexts $l_i$ start with $\Sigma_i^*$
and all the right contexts $r_i$ end with $\Sigma_i^*$, we omit writing
it and assume it by default. The operators are: $\Rightarrow$ for CR
rules, $\Leftarrow$ for SC rules and $\Leftrightarrow$ for composite
rules.

A proposed morphological analysis $P$ is an $n$-tuple of strings, and
the rules are interpreted as applying to a section of this analysis in
context: $P = P_l P_c P_r$ (n-way concatenation of a left context,
centre, and right context). Formally:

\begin{definition}
   A CR rule $(l,c,r)$ {\bf contextually allows}
   $(P_l,P_c,P_r)$ iff $P_l\in l$, $P_r\in r$ and $P_c\in c$.
\end{definition}

\begin{definition}
   An SC rule $(l,c_l,c_r,r)$ {\bf coercively disallows}
   $(P_l,P_c,P_r)$ iff $P_l\in l$, $P_r\in r$, $P_c\in c_l$ and
   $P_c\not\in c_s$.
\end{definition}

\begin{definition}
   A $N$:$M$ {\bf two-level grammar} is a pair
   $(R_{\Rightarrow},R_{\Leftarrow})$, where $R_{\Rightarrow}$ is
   a set of $N$:$M$ context restriction rules and
   $R_{\Leftarrow}$ is a set of $N$:$M$ surface coercion rules.
\end{definition}

\renewenvironment{definition}{\begin{locdef}\em }{\end{locdef}}
\begin{definition}
   \label{def1}
   A two-level grammar $(R_{\Rightarrow},R_{\Leftarrow})$ {\bf
   accepts} the string-tuple $P$, partitioned as $P_1,...,P_k$, iff $P
   = P_1 P_2 ... P_k$ (n-way concatenation) and (1) for each $i$ there
   is a CR rule $A\in R_{\Rightarrow}$ such that $A$
   contextually allows $(P_1...P_{i-1},P_i,P_{i+1}...P_k)$ and (2)
   there are no $ i \leq j$ such that there is an SC rule $B
   \in R_{\Leftarrow}$ such that B coercively disallows
   $(P_1...P_{i-1},P_i...P_{j-1},P_j...P_k)$.
\end{definition}
\renewenvironment{definition}{\begin{locdef}\em }{\hfill$\Box$\end{locdef}}

There are some alternatives to condition (2):

\smallskip(2i)
        there is no $i$ such that there is an SC rule $B \in
         R_{\Leftarrow}$ such that B coercively disallows \break
         $(P_1...P_{i-1},P_i,P_{i+1}...P_k)$: this is (2) with the
         restriction $j=i+1$; since SC rules can only apply to
         the partitions $P_i$, epenthetic rules such as
         $(\Sigma^*\langle
         k,k\rangle,\epsilon\times\Sigma_2^*,\Sigma_1^*\times
         a,\langle k,k\rangle\Sigma^*)$ (`insert an $a$ between
         lexical and surface $k$s') can not be enforced: the rule
         would disallow adjacent $\langle k,k\rangle$s only if they
         were separated by an empty partition: $...\langle
         k,k\rangle,\epsilon,\langle k,k\rangle...$ would be
         disallowed, but $...\langle k,k\rangle,\langle k,k\rangle...$
         would be accepted.

\smallskip(2ii)
         there is no $i$ such that there is an SC rule $B \in
         R_{\Leftarrow}$ such that B coercively disallows \break
         $(P_1...P_{i-1},P_i,P_{i+1}...P_k)$ or B coercively disallows
         $(P_1...P_{i-1},P_i...P_k)$: this is (2) with the restriction
         $j=i+1$ or $j=i$; this allows epenthetic rules to be used but
         may in certain cases be counterintuitive for the user when
         insertion rules are used. For example, the rule
         $(\Sigma^*\langle
         g,g\rangle,u\times\Sigma_2^*,\Sigma_1^*\times v,\Sigma^*)$
         (`change $u$ to $v$ after a $g$') would not disallow a
         string-tuple partitioned as $...\langle
         g,g\rangle,\langle\epsilon,e\rangle,\langle u,u\rangle...$ --
         assuming some CR rule allows $\langle\epsilon,e\rangle$.

\smallskip
Earlier versions of the partition formalism could not (in practice)
cope with multiple lexical characters in SC rules -- see \cite[\S 4.1]{Carter:95}.
This is not the case here.

The following rules illustrate the formalism:
\label{gram}%

\begin{tabular}{l}
      R1:\ \ \ \tlr{V}{b}{*}{$\Rightarrow$}
           {V}{B}{*}{} \\
\end{tabular}

\begin{tabular}{l}
       R2:\ \ \ \tlr{b}{b}{*}{$\Rightarrow$}
           {B}{B}{*}{} \\
\end{tabular}

\begin{tabular}{l}
       R3:\ \ \ \tlr{c}{b}{d}{$\Leftrightarrow$}{c}{}{d}{}
\end{tabular}

R1 and R2 illustrate the iterative application of rules on strings:
they sanction the lexical-surface strings $\langle
\mbox{VBBB},\mbox{Vbbb}\rangle$, where the second $\langle\mbox{B,b}\rangle$
pair serves as the centre of the first application of R2 and as the
left context of the second application of the same rule. R3 is an
epenthetic rule which also demonstrates centres of unequal length.
(We assume that $\langle\mbox{V,V}\rangle$,
$\langle\mbox{c,c}\rangle$ and $\langle\mbox{d,d}\rangle$ are
sanctioned by other identity rules.)

\medskip

The conditions in Definitions~\ref{cr} and~\ref{sc} that restrict the
regular relations in the rules to being `orthogonal' are required in
order for the final language to be regular, because
Definition~\ref{def1} involves an implicit intersection of rule
contexts, and we know that the intersection of regular relations is not
in general regular.

\subsection{Regular Expressions for Compilation}

\newcommand{\beqnhack}{\vspace{-2mm}}
\newcommand{\eeqnhack}{\vspace{-2mm}}

\newcommand{\pad}[1]{\hat{#1}} 
\newcommand{\izeros}[1]{#1^0} 

To compile a two-level grammar into an automaton we use a calculus of
regular languages.  We first use the standard technique of converting
regular n-relations into same-length regular relations by padding them
with a space symbol $0$. Unlike arbitrary regular n-relations,
same-length regular relations are closed under intersection and
complementation, because a theorem tells us that they correspond to
regular languages over ($\epsilon$-free) n-tuples of symbols
\cite[p.~342]{Kaplan:94}.

A proposed morphological analysis $P=P_1...P_k$ can be represented as
a same-length string-tuple $\omega\pad{P_1}\omega\pad{P_2}\omega...\omega\pad{P_k}\omega$,
where $\pad{P_i}\in\Sigma^*$ is $P_i$ converted to a same-length
string-tuple by padding with $0$s, and $\omega=\langle
\omega_1,...,\omega_n\rangle$, where the $\{\omega_i\}$ are new symbols to
indicate the partition boundaries,
$\omega_i\not\in\Sigma_i\cup\{0\}$.

Since in a partitioned string-tuple accepted by the grammar
$(R_{\Rightarrow},R_{\Leftarrow})$ each $P_i\in c$ for some CR
rule $(l,c,r)\in R_{\Rightarrow}$, we can make this representation
unique by defining a canonical way of converting each such possible
centre $C$ into a same-length string-tuple $\pad{C}$. A simple way of doing this
is to pad with $0$s at the right making each string as
long as the longest string in $C$: if $C=\langle p_1,...,p_n\rangle$,\beqnhack
\begin{equation}
\label{padd0}
\pad{C} = \langle p_1 0^*,...,p_n 0^*\rangle\cap\Sigma^*-\Sigma^*\langle 0,...,0\rangle
\eeqnhack
\end{equation}

However, since we know the set of possible partitions
-- it is $\bigcup\{c\mid \exists l,r \langle
l,c,r\rangle\in R_{\Rightarrow}\}$ -- we can reduce the number of
elements of $\Sigma$ in use, and hence simplify the calculations, by
inserting the $0$s in a more flexible manner: e.g., if
$C=\langle ab,b\rangle$, let $\pad{C}=\langle ab,0b\rangle$ rather than
$\pad{C}=\langle ab,b0\rangle$: assuming another rule requires us to use
$\langle b,b\rangle$ anyway, we only have to add $\langle a,0\rangle$
rather than $\langle a,b\rangle$ and $\langle b,0\rangle$. The
preprocessor could use simple heuristics to make such decisions.  In
any case, the padding of possible partitions carries over
to the centres $c$ of CR rules: if $(l,c,r)\in R_{\Rightarrow}$,
$\pad{c}=\{\pad{C}\mid C\in c\}$.
Henceforth let $\pi$ be the set of elements of $\Sigma$ that appear in some
0-padded rule centre.

The 
contexts of all 
rules and the lexical and
surface centres of SC rules must be converted into same-length regular
n-relations by inserting $0$s
at all possible positions
on each tape
independently: if $x=x_1\times...\times x_n$,\beqnhack
\begin{equation}
\label{xbar}
\izeros{x}=({\tt Intro}_{\{0\}} x_1\times...\times{\tt Intro}_{\{0\}}
x_n)\cap\pi^*
\eeqnhack
\end{equation}

Note the difference between this insertion of $0$ everywhere, denoted $\izeros{x}$, and
the canonical padding $\pad{c}$. Both require the
`orthogonality' condition in order for the intersection with $\pi^*$ to
yield a regular language: inserting $0$s into $\langle a,b\rangle^*$ at all possible
positions on each tape independently would give a non-regular relation,
for example.


Now we derive a formula for the set of 0-padded and partitioned
analysis strings accepted by the grammar
$(R_{\Rightarrow},R_{\Leftarrow})$: The set of 0-padded centres
of context restriction rules is given by:\beqnhack
\begin{equation}
   \label{D}
   D=\{\pad{c} \mid \exists l,c,r . (l,c,r) \in R_{\Rightarrow}\}
\eeqnhack
\end{equation}

Here we assume that these centres
are disjoint ($\forall c,d \in D . c = d \vee c \cap d =
\emptyset$), because 
in practice each $c$
is
a singleton set, however there is an alternative derivation
that does not require this.

We proceed subtractively, starting as an initial approximation with an
arbitrary concatenation of the possible partitions, i.e.~the
centres of CR rules:\beqnhack 
\begin{equation}\label{approx1}\displaystyle
\omega(D\omega)^*
\eeqnhack
\end{equation}

From this we wish to subtract the set of strings containing a
partition that is not allowed by any CR rule: We introduce a new
placeholder symbol $\tau$, $\tau\not\in\pi\cup\{\omega\}$, to
represent the centre of a rule, so the set of possible contexts for a
given centre $\pad{c}\in D$ is given by:\beqnhack
\begin{equation}\displaystyle
\bigcup_{(l,\pad{c},r) \in R_{\Rightarrow}}\izeros{l}\tau\izeros{r}
\eeqnhack
\end{equation}

So the set of contexts in which the centre $c$ may {\em not} appear is the
complement of this:\beqnhack
\begin{equation}\displaystyle
\pi^*\tau\pi^*-
\bigcup_{(l,\pad{c},r) \in R_{\Rightarrow}}\izeros{l}\tau\izeros{r}
\eeqnhack
\end{equation}

Now we can introduce the partition separator $\omega$ throughout, then
substitute the centre itself, $\omega \pad{c}\omega$, for its placeholder
$\tau$ in order to derive an expression for the set of partitioned
strings in which an instance of the centre $c$ appears in a context in
which it is {\em not} allowed: [$\circ$ denotes composition]\beqnhack
\begin{equation}\label{Restrict}\displaystyle
{\tt Sub}_{\omega \pad{c}\omega,\tau}\circ{\tt Intro}_{\{\omega\}}\left(
\pi^*\tau\pi^*-
\bigcup_{(l,\pad{c},r) \in R_{\Rightarrow}}\izeros{l}\tau\izeros{r}\right)
\eeqnhack
\end{equation}

If we subtract a term like this for each $\pad{c}\in D$ from our initial
approximation (eq.~\ref{approx1}), then we have an expression for the
set of strings allowed by the CR rules of the grammar:\beqnhack
\begin{eqnarray}\label{approx2}\displaystyle
\omega(D\omega)^*&-& \bigcup_{\pad{c}\in D}
{\tt Sub}_{\omega \pad{c}\omega,\tau}\circ\\
& &{\tt Intro}_{\{\omega\}}\left(
\pi^*\tau\pi^*-
\bigcup_{(l,\pad{c},r) \in R_{\Rightarrow}}\izeros{l}\tau\izeros{r}\right)\nonumber
\eeqnhack
\end{eqnarray}

It remains to enforce the surface coercion rules
$R_{\Leftarrow}$. For a given SC rule $(l,c_l,c_s,r) \in
R_{\Leftarrow}$, a first approximation to the set of strings in
which this rule is violated is given by:\beqnhack
\begin{equation}\displaystyle
{\tt Intro}_{\{\omega\}}(\izeros{l}\omega(\izeros{c_l}-\izeros{c_s})\omega\izeros{r})
\eeqnhack
\end{equation}

Here $(\izeros{c_l}-\izeros{c_s})$ is the set of strings that
match the lexical centre but do not match the surface centre. For part
(2) of Definition~\ref{def1} to apply this must equal the
concatenation of 0 or more adjacent partitions, hence it has on each
side of it the partition separator $\omega$, and the operator {\tt
Intro} introduces additional partition separators into the contexts
and the centre. The only case not yet covered is where the centre
matches 0 adjacent partitions ($i=j$ in part (2) of
Definition~\ref{def1}). This can be dealt with by
prefixing with the substitution operator ${\tt
Sub}_{\omega,\omega\omega}$, so the set of strings in which one of the
SC rules is violated is:\beqnhack
\begin{equation}\label{epenthesis}\displaystyle
\bigcup_{(l,c_l,c_s,r) \in R_{\Leftarrow}}{\tt Sub}_{\omega,\omega\omega}\circ
{\tt Intro}_{\{\omega\}}(\izeros{l}\omega(\izeros{c_l}-\izeros{c_s})\omega\izeros{r})
\eeqnhack
\end{equation}

We subtract this too from our approximation (eq.~\ref{approx2}) in
order to arrive at a formula for the set of 0-padded and partitioned
strings that are accepted by the grammar:\beqnhack
\begin{eqnarray}\label{approx3}\displaystyle
S_0
  &=&\omega(D\omega)^* - \bigcup_{\pad{c}\in D}{\tt Sub}_{\omega \pad{c}\omega,\tau}\circ \nonumber\\
  & & {\tt Intro}_{\{\omega\}}
     \left(\pi^*\tau\pi^*-
     \bigcup_{(l,\pad{c},r) \in R_{\Rightarrow}}\izeros{l}\omega'\izeros{r}\right) \nonumber \\
  &-&\bigcup_{(l,c_l,c_s,r) \in R_{\Leftarrow}}
     {\tt Sub}_{\omega,\omega\omega}\circ \nonumber\\
  & & {\tt Intro}_{\{\omega\}}
     (\izeros{l}\omega(\izeros{c_l}-\izeros{c_s})\omega\izeros{r})
\eeqnhack
\end{eqnarray}

Finally, we can replace the partition separator $\omega$ and the space
symbol $0$ by $\epsilon$ to convert $S_0$ into a regular (but no
longer same-length) relation $S$ that maps between lexical and surface
representations, as in \cite[p.~368]{Kaplan:94}.



\section{Algorithm and Illustration}
\label{illustration}

This section goes through the compilation of the sample grammar in
section~\ref{gram} step by step.  

\subsection{Preprocessing}

Preprocessing involves making all expressions of equal-length.
Let, $\Sigma_1 = \{\mbox{V,B,c,d,0}\}$ and $\Sigma_2 = \{\mbox{V,b,c,d,0}\}$
be the lexical and surface alphabets, respectively.
We pad all centres with 0's (eq.~\ref{padd0}), then compute the set of
0-padded centres (eq.~\ref{D}),\beqnhack
\begin{equation}
D = \{\langle\mbox{B,b}\rangle,%
      \langle\mbox{0,b}\rangle,%
      \langle\mbox{V,V}\rangle,%
      \langle\mbox{c,c}\rangle,%
      \langle\mbox{d,d}\rangle\}
\eeqnhack
\end{equation}
We also compute contexts (eq.~\ref{xbar}). Uninstantiated contexts become\beqnhack
\begin{equation}
 {\tt Intro_{\{0\}}}(\Sigma_1^*)\times
 {\tt Intro_{\{0\}}}(\Sigma_2^*)
\eeqnhack
\end{equation}
The right context of R3, for instance, becomes\beqnhack
\begin{equation}
   {\tt Intro_{\{0\}}}(\mbox{d}\Sigma_1^*)\times
   {\tt Intro_{\{0\}}}(\mbox{d}\Sigma_2^*)
\eeqnhack
\end{equation}


\subsection{Compilation into Automata}

The algorithm consists of three phases:
(1) constructing a
FSA which accepts the centres, (2) applying CR rules, and
(3) forcing SC constraints.

The first approximation to the grammar (eq.~\ref{approx1}) produces
$FSA_1$ which accepts all centres.
\begin{center}
   \ \psfig{figure=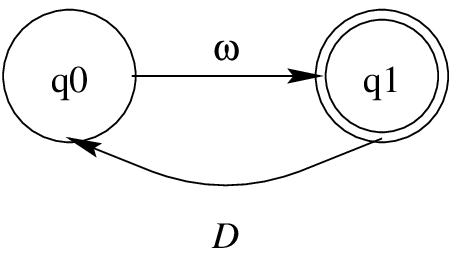,width=3cm}\ \\ $FSA_1$ \\
\end{center}

Phase 2 deals with CR rules. We have two centres to process:
$\langle\mbox{B,b}\rangle$ (R1 \& R2) and $\langle\mbox{0,b}\rangle$
(R3).  For each centre, we compute the set of {\em invalid}
contexts in which the centre occurs (eq.~\ref{Restrict}). Then
we subtract this from $FSA_1$ (eq.~\ref{approx2}), yielding $FSA_2$.
\begin{center}
   \ \psfig{figure=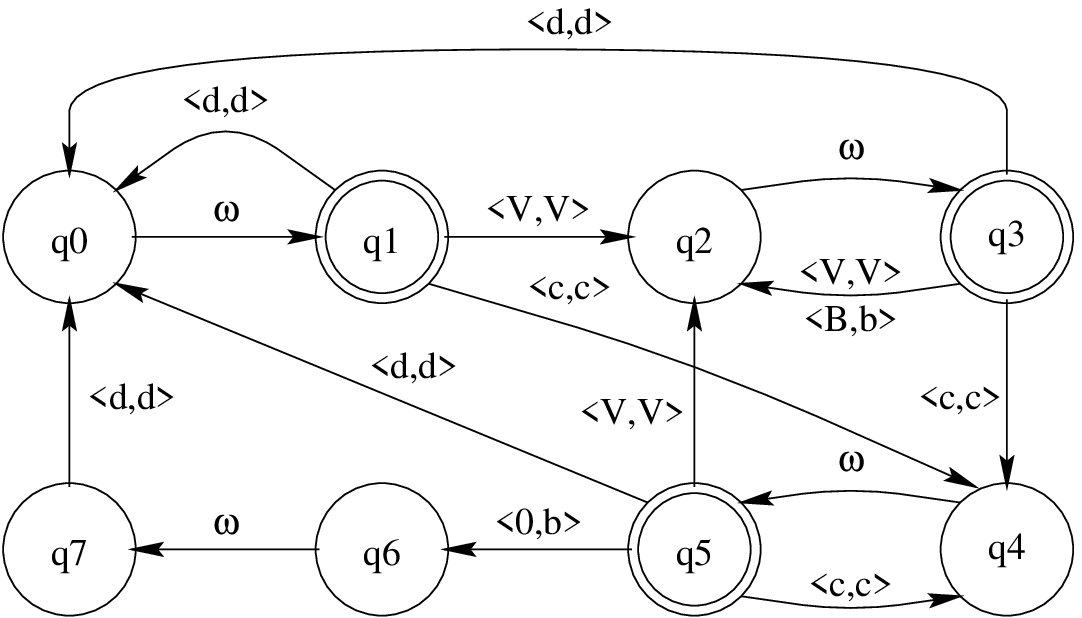,width=7cm}\ \\ 
   $FSA_2$\\
\end{center}

The third phase deals with SC rules: here the $\Leftarrow$ portion of
R3. Firstly, we compute the set of strings in which R3 is violated
(eq.~\ref{epenthesis}). Secondly, we subtract the result from $FSA_2$
(eq.~\ref{approx3}), resulting in an automaton which only differs from
$FSA_2$ in that the edge from $q_5$ to $q_0$ is deleted.

\section{Comparison with Previous Compilations}
\label{comparison}

This section points out the differences in compiling two-level rules
in Koskenniemi's formalism on one hand, and the one presented here on
the other.

\subsection{Overlapping Contexts}

One of the most important requirements of two-level rules is allowing
the multiple applications of a rule on the same string. It is this
requirement which makes the compilation procedures in the Koskenniemi
formalism -- described in \cite{Kaplan:94} -- inconvenient. `The
multiple application of a given rule', they state, `will turn out to
be the major source of difficulty in expressing rewriting rules in
terms of regular relations and finite-state transducers' (p.~346).
The same difficulty applies to two-level rules.

Consider R1 and R2 (\S\ref{gram}), and $D =
\{\langle\mbox{V,V}\rangle,\langle\mbox{B,b}\rangle\}$.
{}\cite{Kaplan:94} express CR rules by the relation,\footnote{This
expression is an expansion of $Restrict$ in \cite[p.~371]{Kaplan:94}.}\beqnhack
\begin{equation}
Restrict(c,l,r) = 
   \ifsthenp{\pi^*l}{c\pi^*}\ \cap\ \ifpthens{\pi^*c}{r\pi^*}
\eeqnhack
\end{equation}
This expression `does not allow for the possibility that the context
substring of one application might overlap with the centre and context
portions of a preceding one' (p.~371). They resolve this by
using auxiliary symbols: (1) They introduce left and right context
brackets, $<_k$ and $>_k$, for each context pair $l_k-r_k$ of a specific
centre which take the place of the contexts. (2) Then
they ensure that each $<_k:<_k$ only
occurs if its context $l_k$ has occurred, and each
$>_k:>_k$ only occurs if followed by its context $r_k$.
The automaton which results after compiling the two rules is:
\begin{center}
     \ \psfig{figure=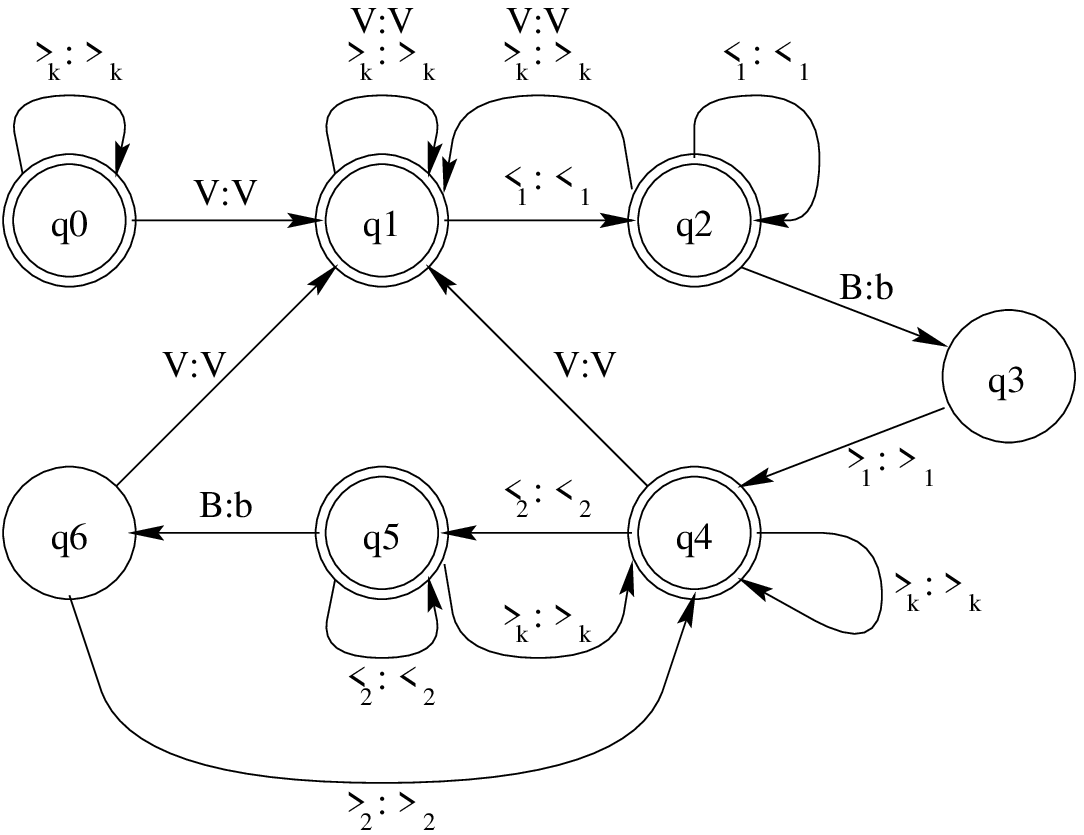,width=7cm}\ 
\end{center}

Removing all auxiliary symbols results in:
\begin{center}
     \ \psfig{figure=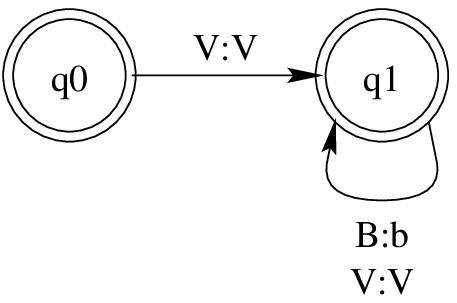,width=3cm}\
\end{center}

Our algorithm produces this machine directly.
Compiling Koskenniemi's formalism is complicated
by its interpretation: rules apply to the
entire input. A \IT\ rule is concerned only
with the part of the input that matches its centre.

\subsection{Conditional Compilation}

Compiling epenthetic rules in the Koskenniemi formalism requires
special means; hence, the algorithm is conditional on the type of the
rule \cite[p.~374]{Kaplan:94}. This peculiarity, in the Koskenniemi
formalism, is due to the dual interpretation of the {\tt 0} symbol in
the parallel formalism: it is a genuine symbol in the alphabet, yet it
acts as the empty string $\epsilon$ in two-level expressions. Note
that it is the duty of the user to insert such symbols as
appropriate \cite{Karttunen:92}.

This duality does not hold in the \IT formalism.  The user can express
lexical-surface pairs of unequal lengths. It is the duty of the rule
compiler to ensure that all expressions are of equal length prior to
compilation. With CR rules, this is done by padding
zeros. With SC rules, however, the {\tt Intro} operator
accomplishes this task.  There is a subtle, but important, difference
here.

Consider rule R3 (\S\ref{gram}).  The 0-padded centre of the CR
portion becomes \lab 0,b\rab.  The SC portion, however, is
computed by the expression\beqnhack
\begin{equation}
   {\tt Insert}_{\{0\}}(\epsilon)\times
   \overline{{\tt Insert}_{\{0\}}(b)}
\eeqnhack
\end{equation}
yielding automaton (a):

\begin{tabular}{cc}
   \psfig{figure=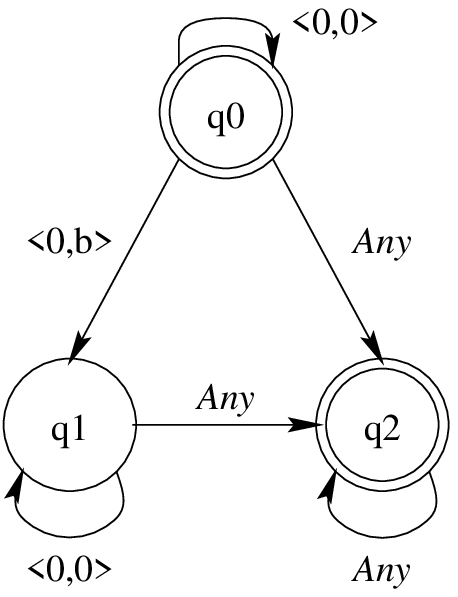,width=3cm}\ \ \ &
   \psfig{figure=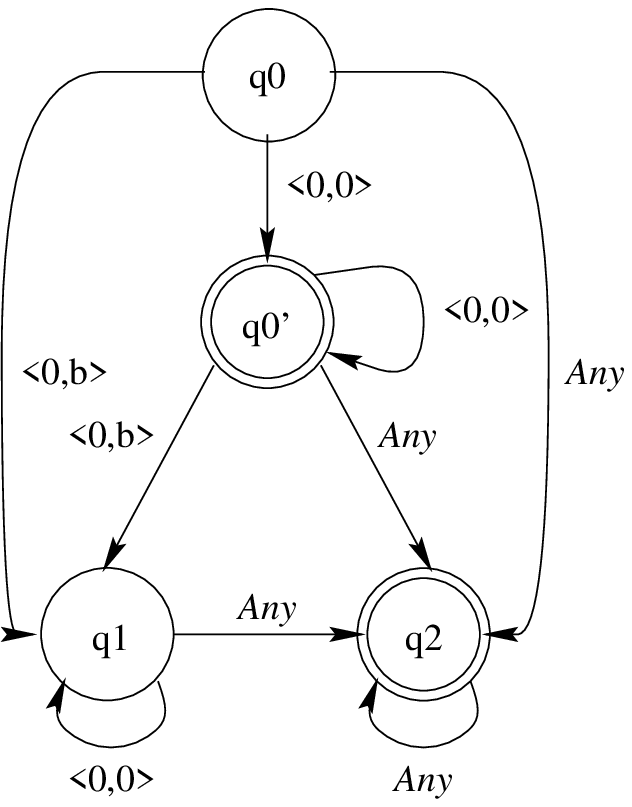,width=3.5cm}\\
   a & b
\end{tabular}

If the centre of the SC portion had been padded with 0's,
the centre would have been\beqnhack
\begin{equation}
   {\tt Insert}_{\{0\}}(0)\times
   \overline{{\tt Insert}_{\{0\}}(b)}
\eeqnhack
\end{equation}
yielding the undesired automaton (b). Both are
similar except that state $q_0$ is final in the former.  Taking
(a) as the centre, eq.~\ref{epenthesis} includes \lab
cd,cd\rab; hence, eq.~\ref{approx3} excludes it.  The compilation of our
rules is not conditional; it is general enough to cope with all sorts
of rules, epenthetic or not.

\section{Conclusion and Future Work}
\label{conclusion}

This paper showed how to compile the \IT formalism into N-tape
automata. Apart from increased efficiency and portability of
implementations, this result also enables us to more easily relate
this formalism to others in the field, using the finite-state calculus
to describe the relations implemented by the rule compiler.

A small-scale prototype of the algorithm has been implemented in
Prolog.  The rule compiler makes use of a finite-state calculus
library which allows the user to compile regular expressions into
automata. The regular expression language includes standard operators
in addition to the operators defined here. The system has been tested
with a number of hypothetical rule sets (to test the integrity of the
algorithm) and linguistically motivated morphological grammars which
make use of multiple tapes. Compiling realistic descriptions would
need a more efficient implementation in a more suitable language such
as C/C++.


Future work includes an extension
to simulate
a restricted form of  unification between  categories associated
with rules and morphemes.
\vspace{-2mm}

\footnotesize
\newcommand{\bibhack}{\vspace{-1.5mm}}

\end{document}